\documentclass[conference]{IEEEtran}
\IEEEoverridecommandlockouts
\usepackage{booktabs} 
\usepackage{graphicx}
\usepackage{setspace}
\usepackage{amsmath}
\usepackage{dsfont}
\usepackage{amsfonts}
\usepackage{lipsum}
\usepackage{multicol}
\usepackage{float}
\usepackage{bm}
\usepackage{color}
\usepackage[dvipsnames]{xcolor}
\usepackage{etoolbox}
\usepackage{soul}
\usepackage{amssymb}
\usepackage[linesnumbered,ruled,vlined]{algorithm2e}
\usepackage{mathtools,amssymb}
\usepackage{caption}
\usepackage{float}
\usepackage{amsthm}
\usepackage[caption = false,subrefformat=parens,labelformat=parens]{subfig}
\usepackage{adjustbox}
\usepackage{afterpage}
\usepackage{multirow}
\usepackage{mathrsfs}
\usepackage{hyperref}
\usepackage{makecell}

\def\BibTeX{{\rm B\kern-.05em{\sc i\kern-.025em b}\kern-.08em
    T\kern-.1667em\lower.7ex\hbox{E}\kern-.125emX}}

\begin{document}

\title{TQ-DiT: Efficient Time-Aware Quantization for Diffusion Transformers

\author{
Younghye Hwang, Hyojin Lee, Joonhyuk Kang \\
School of Electrical Engineering, KAIST, South Korea\\
Email: yhwang@kaist.ac.kr, jkang@kaist.ac.kr
}

\thanks{This research was supported by the MSIT (Ministry of Science and ICT), Korea, under the ITRC (Information Technology Research Center) support program (IITP-2025-0-01787) supervised by the IITP (Institute of Information \& Communications Technology Planning \& Evaluation).\\
}
}

\maketitle

\begin{abstract}

Diffusion transformers (DiTs) combine transformer architectures with diffusion models. However, their computational complexity imposes significant limitations on real-time applications and sustainability of AI systems. In this study, we aim to enhance the computational efficiency through model quantization, which represents the weights and activation values with lower precision. Multi-region quantization (MRQ) is introduced to address the asymmetric distribution of network values in DiT blocks by allocating two scaling parameters to sub-regions. Additionally,  time-grouping quantization (TGQ) is proposed to reduce quantization error caused by temporal variation in activations. The experimental results show that the proposed algorithm achieves performance comparable to the original full-precision model with only a 0.29 increase in FID at W8A8. Furthermore, it outperforms other baselines at W6A6, thereby confirming its suitability for low-bit quantization. These results highlight the potential of our method to enable efficient real-time generative models.

\end{abstract}

\section{Introduction}
\label{sec:intro}

Diffusion models have emerged as a promising method for generative tasks due to their stable image generation by iteratively refining noise into structured data \cite{DMsurvey2}. 
Initially, diffusion models employed U-Net architectures to capture local patterns and hierarchical features \cite{ADM, LDM}. 
However, several works have revealed that the U-Net structure is unnecessary and limits task flexibility and scalability \cite{ViT, Transformer1}. 
Furthermore, diffusion transformers (DiTs) have been introduced by exploiting the transformer architecture, which has task flexibility and scalability, for the backbone networks in diffusion models \cite{DiT, DiT2, DiT3, DiT4}.

However, DiTs require substantial computational resources due to their large number of parameters and iterative sampling process. 
For example, generating a $256\times256$ image with DiT-XL-2 over 1000 timesteps takes approximately 14 seconds on an NVIDIA RTX 4090. 
Even sampling a $512\times512$ image with 256 timesteps requires 50 TFLOPS and takes over 15 seconds on the same GPU.
The extended sampling duration hampers real-world deployment, particularly in resource-limited environments. 
Furthermore, the extremely high resource consumption undermines efforts toward sustainable artificial intelligence (AI).

For sustainable AI, model compression can be employed to reduce both computational loads and GPU dependency \cite{SusAI1, SusAI2}.
Model compression techniques, including quantization, pruning, and knowledge distillation, aim to reduce computational cost and model size in deep neural networks. Among these, quantization is particularly efficient, converting floating-point weights and activations into discrete integer values without modifying the model architecture. In \cite{Qinference}, for instance, 8-bit quantized models on ARM CPUs achieve 2.2× faster inference than their floating-point counterparts.

\begin{figure}
    \centering
    \includegraphics[width=\columnwidth]{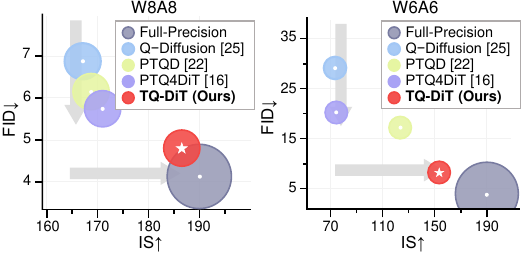}
    \caption{Quantization performance is examined with weights and activations at both 8-bit precision (W8A8) and 6-bit precision (W6A6). The proposed TQ-DiT scheme achieves performance closest to that of the original full-precision models, as observed by the lowest FID and highest IS among the conventional quantization schemes.
    } \label{fig1}
    \vspace{0.5cm}
\end{figure}

Post-training quantization (PTQ), widely studied in classification and object detection \cite{PTQ1,PTQ2,PTQ3}, offers a practical solution for reducing inference time requiring only a small calibration dataset without retraining. Unlike quantization-aware training (QAT), which requires extensive fine-tuning and high GPU usage, PTQ is a more suitable option for large-scale models that demand substantial computational resources \cite{PTQ4DiT}. PTQ’s efficiency is particularly crucial for edge servers in distributed systems, as lightweight quantization minimizes calibration overhead and reduces energy consumption during inference \cite{energy}. Furthermore, quantizing DiTs enables efficient semantic communication \cite{Semantic} by facilitating rapid image generation in latency-sensitive scenarios \cite{Semantic2}.

\begin{figure}
    \centering
    \includegraphics[width=\columnwidth]{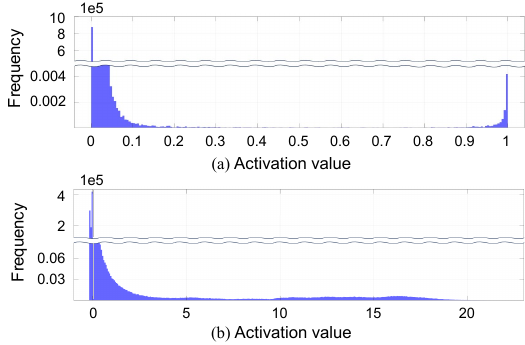}
    \caption{Distribution of values after the Softmax (a), GELU (b) in DiT blocks. 
    Since the values are non-uniformly distributed, conventional quantization can degrade performance significantly.
    } \label{fig2}
\end{figure}

\begin{figure}
    \centering
    \includegraphics[width=\columnwidth]{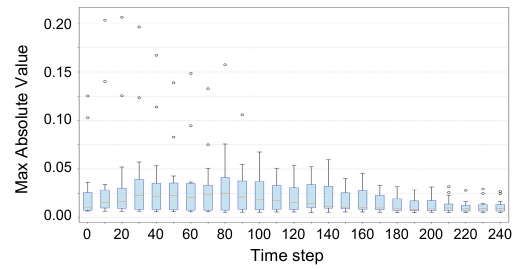}
    \caption{Maximum channel magnitudes after softmax are depicted for various timesteps during inference, revealing large variance across timesteps. This shows the necessity of handling timestep-dependent values effectively.
    } \label{fig3}
\end{figure}

The iterative sampling process and transformer architecture of DiTs introduce two main challenges for PTQ.
i) \textit{Asymmetric activation distribution in DiTs.} As illustrated in Fig. \ref{fig2}, post-softmax activations are concentrated near zero within the [0,1] range, whereas post-GELU activations exhibit a wider spread and a negative skew. These differences make it difficult to accurately capture both distributions using a single quantization parameter.
ii) \textit{Activation distribution variation across timesteps.} The activation distribution varies across multiple timesteps during inference. As DiTs generate images from random noise through an iterative diffusion process, the maximum absolute value of post-softmax activations fluctuates significantly, as illustrated in Fig. \ref{fig3}. Quantization parameters optimized for a specific timestep may not generalize well to others, leading to significant errors. This discrepancy causes deviations from the original full-precision distributions and degrades the performance of the quantized model.
    
To address these challenges, we propose an efficient time-aware quantization algorithm for diffusion transformers (TQ-DiT) within a PTQ framework. The contributions are summarized as follows:
\begin{itemize}
    \item We introduce a simple yet efficient time-grouping quantization method to address the challenge of time variance in DiT quantization, achieving significant performance improvements with minimal memory overhead.
    \item Our approach achieves performance comparable to the original full-precision model at 8-bit precision and surpasses other baselines at 6-bit precision, striking a balance between efficiency and accuracy.
    \item TQ-DiT reduces GPU resource requirements for quantization, supporting sustainable AI initiatives, while generating higher-quality images using smaller calibration datasets compared to prior methods.
\end{itemize}

The remainder of this paper is organized as follows. Section \ref{sec:background} reviews the concepts of diffusion models, DiTs, and model quantization. In Section \ref{sec:method}, we investigate the asymmetric distributions of weights and activations, and propose TQ-DiT, which addresses the asymmetries for DiTs. Section \ref{sec:experiments} presents the numerical results comparing the performance of TQ-DiT with benchmarks. Finally, Section \ref{sec:conclusion} concludes the findings and discusses future directions.

\section{Background and Related works}
\label{sec:background}

\subsection{Diffusion Models}
\label{ssec:DM}
Diffusion models operate by gradually introducing Gaussian noise to real data $x_0$ in a forward process and then learning a reverse process to denoise and generate high-quality images. For denoising diffusion probabilistic models (DDPMs) \cite{DM}, the forward process is defined as a Markov chain, represented by the following equation:

\begin{equation}
q(x_t | x_{t-1}) = \mathcal{N}(x_t; \sqrt{\alpha_t} x_{t-1}, \beta_t I),
\end{equation}
where $\alpha_t$ and $\beta_t$ are hyperparameters, with $\beta_t = 1 - \alpha_t$. 

In the reverse process, since directly modeling the true distribution $q(x_{t-1}|x_t)$ is infeasible, diffusion models employ variational inference to approximate it as a Gaussian distribution:

\begin{equation}
p_\theta(x_{t-1}|x_t) = \mathcal{N}(x_{t-1}; \mu_\theta(x_t, t), \Sigma_\theta(x_t, t)).
\end{equation}
The mean of the Gaussian can further be reparameterized using a noise prediction network $\epsilon_\theta(x_t, t)$ as follows.

\begin{equation}
\mu_\theta(x_t, t) = \frac{1}{\sqrt{\alpha_t}} \left( x_t - \frac{1 - \alpha_t}{\sqrt{1 - \bar{\alpha}_t}} \epsilon_\theta(x_t, t) \right),
\end{equation}
where $\bar{\alpha}_t = \prod_{s=1}^t \alpha_s$. The variance $\Sigma_\theta(x_t, t)$ can either be reparameterized or set to a fixed schedule $\sigma_t$. Under the fixed variance schedule, the distribution of $x_{t-1}$ is given by

\begin{equation}
x_{t-1} \sim \mathcal{N}(x_{t-1}; \mu_\theta(x_t, t), \sigma_t^2 I).
\end{equation}

In diffusion models, the noise at each time step $t$ is predicted from $x_t$ using a noise estimation model, which typically shares the same weights across all time steps.

\begin{figure}
    \centering
    \includegraphics[width=\columnwidth]{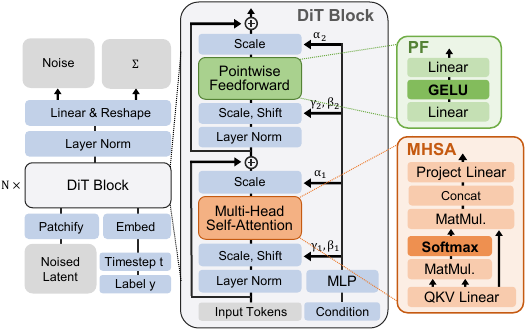}
    \caption{Illustration of the diffusion transformer (DiT)~\cite{DiT} with stacked transformer-based DiT blocks. Each block includes MHSA layers with softmax and PF layers with GELU activations, conditioned on class and timestep inputs.
    } \label{fig4}
\end{figure}

\subsection{Diffusion Transformers}
\label{ssec:DiT}
Despite considerable impact of the U-Net architecture on image generation models \cite{DM, ADM, LDM}, recent studies have shifted toward transformer-based approaches \cite{DMsurvey2}. Diffusion transformers (DiTs)~\cite{DiT} show state-of-the-art performance in image generation, while they can scale effectively in terms of data representation and model size.

DiTs are structured with $N$ transformer-based blocks that form the backbone of the denoising process, as depicted in Fig. \ref{fig4}. Each block includes two fundamental components: multi-head self-attention (MHSA) and pointwise feedforward (PF) layers. Both components are conditioned on class information and timestep inputs, ensuring the model effectively captures time-dependent features throughout the denoising process. 

MHSA mechanism primarily relies on linear projections and matrix multiplications (MatMul) of the query, key, and value matrices, allowing the model to capture contextual relationships among image patches. Each DiT block employs a softmax layer in the MHSA to normalize attention scores and effectively capture relative importance among tokens. This normalization is critical for the self-attention mechanism to function properly. For PF layers, two sequential linear transformations are applied, separated by a Gaussian Error Linear Unit (GELU) activation layer.

Although DiTs have shown remarkable efficiency in generating high-fidelity images, their significant computational demands present challenges for practical applications. To address this limitation, we propose a quantization framework designed for DiTs, substantially reducing memory usage and inference time. Notably, our approach achieves this efficiency without requiring re-training of the original model, making it a practical and scalable solution for deploying DiTs in resource-constrained environments.

\begin{figure*}
    \centering
    \includegraphics[width=\textwidth]{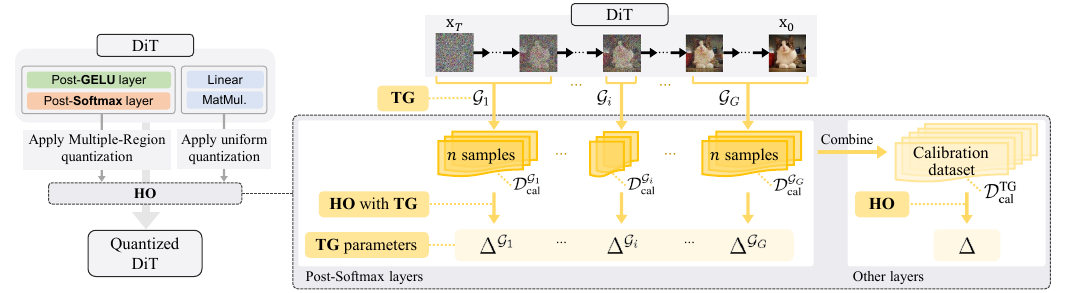}
    \caption{Illustration of the proposed TQ-DiT. Multi-Region Quantization (MRQ) handles skewed distributions in post-softmax and post-GELU layers within MHSA and PF. Hessian-guided Optimization (HO) with Time-Grouping Quantization(TGQ) addresses timestep-dependent activation variability in post-softmax layers.
    } \label{fig6}
\end{figure*}

\subsection{Model Quantization}
\label{ssec:DiT}

Quantization is employed for model compression to enhance the inference efficiency of deep learning models by converting full-precision tensors into $k$-bit integer representations \cite{PTQ4}. This conversion leads to significant improvements in computational efficiency and reductions in memory usage \cite{Qinference}.

For uniform quantization, the process can be mathematically expressed as

\begin{equation}
\hat{\mathbf{x}} = s \cdot \text{clip} \left( \lfloor\frac{\mathbf{x}}{s} \rceil + z, 0, 2^k-1 \right) - z,
\end{equation}
where $\lfloor \cdot \rceil$ denotes the rounding operation, $s = \frac{\max(\mathbf{x}) - \min(\mathbf{x})}{2^k-1}$ is the step size, and $z = -\lfloor \frac{\min(\mathbf{x})}{s} \rceil$ is the zero-point. Here, $k$ represents the bit-width of the quantization. This formula essentially maps floating-point values to a predefined set of fixed points (or grids).

For $k$-bit uniform asymmetric quantization, the set of quantization grids can be expressed as

\begin{equation}
\mathcal{Q}_k^\text{u} = s \times \{0, \dots, 2^k-1\} - z.
\end{equation}

The quantization function, denoted as $Q_k(\cdot \,;\Delta):\mathbb{R} \to \mathcal{Q}_k^\text{u}$, is often optimized to minimize the quantization error, defined by the deviation between the original and quantized grids. The optimization is formulated as

\begin{equation}
\min_{s, z} ||\hat{\mathbf{w}} - \mathbf{w}||_F^2 \quad \text{s.t.} \quad \hat{\mathbf{w}} \in \mathcal{Q}_k^\text{u},
\end{equation}
where $\mathbf{w}$ is the original parameter, and $\hat{\mathbf{w}}$ represents its quantized counterpart.

However, recent studies obtain that merely minimizing the quantization error in the parameter space does not always yield optimal task performance. Instead, task-aware approaches focus on minimizing the final task-specific loss function, such as cross-entropy or mean squared error, with quantized parameters. The task-aware approach can be expressed as

\begin{equation}
\label{quantization obj}
\min_{\Delta} \mathbb{E}[\mathcal{L}(\hat{\mathbf{w}})] \quad \text{s.t.} \quad \hat{\mathbf{w}} \in \mathcal{Q}_k^\text{u},
\end{equation}
where $\mathcal{L}(\cdot)$ denotes the task-specific loss function and $\Delta=\{s,z\}$ is quantization parameters. The task-aware quantization has demonstrated better preservation of model performance compared to conventional methods.

Among the various quantization techniques, PTQ has become popular for large-scale models due to its efficiency and ability to avoid resource-intensive re-training \cite{PTQ1, PTQD, PTQ4DiT}. PTQ utilizes a small calibration dataset to fine-tune the quantization parameters, enabling quantized models to achieve performance close to full-precision counterparts with minimal data and computation. It has been successfully applied to diverse architectures, including CNNs \cite{PTQ1, PTQ2}, language transformers \cite{QLLM, QLLM2}, vision transformers (ViTs) \cite{PTQ3}, and U-Net-based diffusion models \cite{PTQD, Q-Diffusion}.

A recent study extended PTQ to DiTs, introducing a technique that redistributes activations and weights based on their salience to mitigate quantization errors caused by outlier magnitudes \cite{PTQ4DiT}. However, this approach is limited by its reliance on salience-based redistribution, which requires extensive calibration time and a large-scale calibration dataset, imposing significant computational and resource burdens. Such inefficiencies are particularly problematic in real-world applications with limited computational resources, such as edge servers in distributed systems \cite{chainnet_edge_ai}, where efficient quantization strategies are critical for deployment. In comparison, our work proposes an alternative PTQ strategy that directly targets quantization errors across DiT, significantly reducing calibration overhead while maintaining generation quality. By addressing these inefficiencies, our study introduces a streamlined approach for effectively quantizing DiTs, enabling their deployment in resource-constrained environments without compromising performance in high-quality image generation tasks.

\section{Methodology}
\label{sec:method}
The proposed TQ-DiT framework quantizes diffusion transformers (DiTs) through a structured approach that integrates three components: Time-Grouping Quantization (TGQ), Hessian-Guided Optimization (HO), and Multi-Region Quantization (MRQ). The framework is designed to address timestep-dependent variance, parameter sensitivity, and non-uniform activation distributions in DiTs.

\subsection{Time-Grouping Quantization}
\label{ssec:TG}
The timestep-dependent variance of the values introduces inconsistency in quantization. 
To address this challenge, we propose a time-grouping quantization (TGQ) algorithm, which effectively manages variance by grouping timesteps and optimizing quantization parameters for each group.

\subsubsection{Calibration Dataset Construction with Time Grouping}
imesteps $\{0, 1, \dots, T-1\}$ are divided into $G$ groups, where each group represents a contiguous segment of timesteps as follows.

\begin{equation}
\mathcal{G}_i = \left\{ t \ \middle| \ t \in \left[ \frac{(i-1)T}{G}, \frac{iT}{G} - 1 \right] \right\}, \quad \forall i \in \{1, \dots, G\}.
\end{equation}
From each group $\mathcal{G}_i$, $n$ samples for calibration are randomly selected to capture a balanced representation of timestep-specific activation distributions. The total calibration dataset is given by
\begin{equation}
\mathcal{D}_{\text{cal}}^{\text{TG}} = \bigcup_{i=1}^{G} \mathcal{D}_{\text{cal}}^{\mathcal{G}_i}, \quad \text{where } |\mathcal{D}_{\text{cal}}^{\text{TG}}| = n \cdot G.
\end{equation}
\noindent
This provides sufficient data diversity to inform the optimization of timestep-specific quantization parameters.

\subsubsection{Time-Grouping Optimization}
Quantization inherently introduces a trade-off between parameter precision and task performance. As previously defined in \eqref{quantization obj}, the objective for quantization is to minimize the task loss $\mathcal{L}$. In the context of DiTs, the task loss $\mathcal{L}$ is further specified as
\begin{equation}
\label{DM loss}
\mathcal{L} = \mathbb{E}_{x_0, t, \epsilon \sim \mathcal{N}(0, I)} \left[ \left\| \epsilon - \epsilon_\theta(x_t, t) \right\|_2^2 \right],
\end{equation}
where $x_t$ represents the noisy data at timestep $t$, $\epsilon$ denotes the added noise sampled from a standard Gaussian distribution, and $\epsilon_\theta(x_t, t)$ is the predicted noise. For each timestep group, quantization parameters for activation $A$, $\Delta_A^{l, \mathcal{G}_i}$, are optimized to reduce quantization errors, defined as

\begin{equation}
\label{group obj}
\Delta_A^{l, \mathcal{G}_i} = \arg \min_{\Delta} \mathbb{E} \left[ \left\| \epsilon_{\hat{\theta}}^l(x_t, t; \Delta) - \epsilon_{\theta}^l(x_t, t) \right\|^2 \right],
\end{equation}
where $\epsilon_{\theta}^l(x_t, t)$ is the activation at layer $l$, $\epsilon_{\hat{\theta}}^l(x_t, t; \Delta)$ is the quantized activation, and $\Delta$ represents the quantization parameters containing step size $s$ and zero point $z$. 
When a layer has a sensitive activation distribution, such as post-softmax, the time-grouping optimization enhances performance by mitigating quantization errors caused by temporal variations. 
Since activation distributions change across timesteps, using a single quantization parameter for all timesteps fails to capture these shifts. 
By assigning separate parameters to grouped timesteps, the quantization process better preserves distribution characteristics, reducing distortion and improving accuracy.

\subsection{Hessian-Guided Optimization with Time Grouping}

To simultaneously achieve high accuracy and efficient quantization, we propose to utilize Hessian-guided optimization (HO) for DiTs. Unlike mean square error (MSE) or cosine distance, which focus only on numerical or geometric similarity \cite{MSE1, MSE2}, HO incorporates squared gradient information, emphasizing outputs with higher absolute gradients due to their stronger influence on the loss \cite{PTQ1}. 

When weights are considered as variables, the quantization loss is defined as the difference between the original task loss and the loss introduced by quantization perturbations. We can approximate the expected quantization loss using a Taylor series expansion as follows \cite{PTQ2}.
\begin{equation}
\mathbb{E}[\mathcal{L}(\mathbf{\theta} + \Delta\mathbf{\theta})-\mathcal{L}(\mathbf{\theta} )] 
\approx \Delta\mathbf{\theta}^\mathsf{T} \bar{\mathbf{g}}^{(\mathbf{\theta})} + \frac{1}{2} \Delta\mathbf{\theta}^\mathsf{T} \bar{\mathbf{H}}^{(\mathbf{\theta})} \Delta\mathbf{\theta},
\end{equation}
where $\theta=[\mathbf{w}^{(1),\mathsf{T}},...,\mathbf{w}^{(L),\mathsf{T}}]^{\mathsf{T}}$ is the stacked vector of weights in all $L$ layers, $\bar{\mathbf{g}}^{(\mathbf{\theta})}=\mathbb{E}[\nabla_\mathbf{\theta} \mathcal{L}]$ and $\bar{\mathbf{H}}^{(\mathbf{\theta})}=\mathbb{E}[\nabla_\mathbf{\theta}^2 \mathcal{L}]$ are gradients and the Hessian matrix, and $\Delta\theta$ denotes the quantization-induced perturbation. Assuming the model is trained to convergence, the gradient term is negligible. The optimization objective then reduces to minimize the second-order term $\Delta\theta^\mathsf{T} \bar{\mathbf{H}}^{(\theta)} \Delta\theta$.

Given the pre-activation outputs of layer $l$, which are the values preceding the activation layer and denoted as $\mathbf{z}^{(l)}$, the optimization problem can be reformulated as
\begin{equation}
\min \mathbb{E} \left[ \Delta\mathbf{z}^{(l),\mathsf{T}} \mathbf{H}^{(\mathbf{z}^{(l)})} \Delta\mathbf{z}^{(l)} \right],
\end{equation}
where $\Delta\mathbf{z}^{(l)}$ is the difference between full-precision and quantized pre-activation outputs, and $\mathbf{H}^{(\mathbf{z}^{(l)})}$ is the pre-activation Hessian matrix. The pre-activation Hessian is approximated by using the diagonal Fisher information matrix (FIM) \cite{PTQ1}. Thus, the objective can be expressed as
\begin{equation}
\min \mathbb{E} \left[ 
\Delta\mathbf{z}^{(l),\mathsf{T}} 
\text{diag}\left( \left( \frac{\partial \mathcal{L}}{\partial \mathbf{z}^{(l)}_1} \right)^2, \cdots, \left( \frac{\partial \mathcal{L}}{\partial \mathbf{z}^{(l)}_a} \right)^2 \right) 
\Delta\mathbf{z}^{(l)} 
\right].
\end{equation}

In the case of DiTs, where the output $\epsilon_\theta(x_t, t)$ represents predicted noise at timestep $t$ and the task loss $\mathcal{L}$ is defined as \eqref{DM loss}, the optimization problem for layer $l$ is given by
\begin{align}
\min \mathbb{E} \Bigg[ &
\Delta\epsilon^{(l)}(x_t, t)^\mathsf{T} \mathbf{G}^{(l)} \Delta\epsilon^{(l)}(x_t, t) \Bigg],
\label{final_obj}
\end{align}
where \(\Delta\epsilon^{(l)}(x_t, t)\) denotes the difference between full-precision and quantized pre-activation noise at \(l\)-th layer in DiT, defined as \(\Delta\epsilon^{(l)}(x_t, t) = \epsilon_\theta^{(l)}(x_t, t) - \epsilon_{\hat{\theta}}^{(l)}(x_t, t; \Delta)\), and $\mathbf{G}^{(l)} = \text{diag}\big(\big(\frac{\partial \mathcal{L}}{\partial \epsilon_\theta^{(l)}}\big)^2\big).$

HO is integrated with TGQ algorithm. The TGQ objective function in \eqref{group_obj} is reformulated as
\begin{align}
\Delta_A^{l, \mathcal{G}i} = \arg \min{\Delta} \mathbb{E}_{t \in \mathcal{G}_i} \Bigg[ &
\Delta \epsilon^{(l)} (x_t, t)^\mathsf{T} \mathbf{G}^{(l)} \Delta \epsilon^{(l)} (x_t, t) \Bigg].
\label{group_final_obj}
\end{align}
This combined approach leverages both parameter sensitivity and temporal dynamics, resulting in improved quantization performance for generative tasks.

\begin{algorithm}[t]
\DontPrintSemicolon

\textbf{Input:} Pre-trained DiT $\mathcal{M}(\theta)=f^{L}_{\theta^{L}} \circ f^{L-1}_{\theta^{L-1}} \circ \dotsb \circ f^{1}_{\theta^{1}}$, timesteps $T$, target bit $k$, groups $G$, calibration samples $n$\;
\textbf{Output:} Quantized DiT $\mathcal{M}(\hat{\theta}_k)$\;

\textbf{Phase 1: Calibration Dataset Generation}\;
\For{each timestep $t \in \{0, 1, \dots, T-1\}$}{
    Collect tuples $(x_t, t, y)$\;
}
\For{$i = 1$ \KwTo $G$}{
    Divide timesteps into $G$ groups, $\mathcal{G}_i = \{ t \mid t \in [(i-1)T/G, iT/G - 1] \}$\;
    Randomly sample $n$ tuples $\mathcal{D}_{\text{cal}}^{\mathcal{G}_i} = \{(x_t, t, y) \mid t \in \mathcal{G}_i\}$\;
}

\textbf{Phase 2: Layer-Wise Computation}\;
\For{each sample $(x_t, t, y) \in \mathcal{D}_{\text{cal}}^{\text{TG}}$}{
    Perform FP to compute $\epsilon_\theta^l(x_t, t)$ for $l = 1, \dots, L$\;
    Perform BP to compute $\nabla_{\epsilon_{\theta}} \mathcal{L}$ for $l = L, \dots, 1$\;
}

\textbf{Phase 3: Time-Aware Quantization}\;
\For{$l = 1$ \KwTo $L$}{
    \If{layer $l$ is CNN or linear layer}{
    Generate candidates for $\Delta_W^l$ and $\Delta_X^l$\;
    \For{iteration $r = 1$ \KwTo $R$}{
        Update $\Delta_W^l$ with \textbf{HO}\;
        \If{layer $l$ is the \textbf{Post-GELU} layer}{
            Update $\Delta_X^l$ using \textbf{MRQ} with \textbf{HO}\;
        \Else{
            Update $\Delta_X^l$ with \textbf{HO}\;
            }
        }
    }}
    \ElseIf{layer $l$ is matrix-multiplication layer}{
    Generate candidates for $\Delta_A^l$ and $\Delta_B^l$\;
    \For{iteration $r = 1$ \KwTo $R$}{
        \tcc{Time-Grouping quantization}
        \If{layer $l$ is the \textbf{Post-softmax} layer}{
            \For{each group $\mathcal{G}_i$}{
                Update $\Delta_A^{l, \mathcal{G}_i}$ using \textbf{MRQ} with \textbf{HO}\;
            }
        \Else{
            Update $\Delta_A^l$ with \textbf{HO}\;
            }
        }
        Update $\Delta_B^l$ with \textbf{HO}\;
    }
    }
}

\caption{Time-Aware Quantization}
\label{alg}
\end{algorithm}

\begin{figure*}
    \centering
    \includegraphics[width=\textwidth]{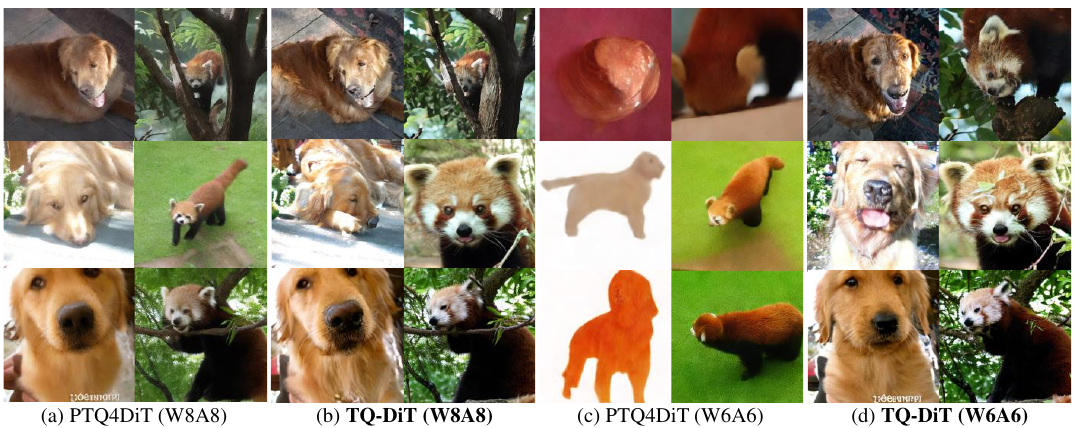}
    \caption{Random samples generated by TQ-DiT for W8A8 (b), W6A6 (d) and a strong baseline, PTQ4DiT~\cite{PTQ4DiT}, for W8A8 (a) and W6A6 (c) on ImageNet 256x256. TQ-DiT produces sharper results under W8A8 and preserves fine details better under W6A6 compared to PTQ4DiT~\cite{PTQ4DiT}.
    } \label{fig5}
\end{figure*}

\subsection{Multi-Region Quantization}
\label{ssec:MR}
Multi-region quantization (MRQ) technique is adapted for non-uniformly distributed values of DiT blocks.
For softmax activations, the range is divided into two regions: \( R_1 = [0, 2^{k-1}s_1) \) and \( R_2 = [2^{k-1}s_1, 1] \). 
Small values in \( R_1 \) are quantized using step size \( s_1 \), while larger values in \( R_2 \) employs a fixed step size \( s_2 = \frac{1}{2^{k-1}} \). The optimal \( s_1 \) is determined by the objective function of TGQ, as defined in \eqref{group_final_obj}, where \(\Delta\) includes \( s_1 \). Similarly, for GELU activations, the asymmetric distribution is partitioned into \( R_1 = [-2^{k-1}s_1^g, 0] \) and \( R_2 = [0, 2^{k-1}s_2^g) \), with separate step sizes \( s_1^g \) and \( s_2^g \) calibrated independently for positive and negative values to minimize quantization error.
This method, initially validated for classification tasks in ViTs \cite{PTQ3}, is combined with HO based TGQ and experimentally shown to be equally effective for image generation tasks in DiTs.

\subsection{TQ-DiT Framework}
The proposed framework quantizes DiTs through three sequential stages, detailed as follows:

\begin{itemize}
    \item \textbf{Phase 1: Calibration Dataset Generation}:
    To account for timestep variability, the timesteps \(\{0, 1, \dots, T-1\}\) are divided into \(G\) groups \(\{\mathcal{G}_1, \dots, \mathcal{G}_G\}\). From each group, \(n\) samples are randomly selected to build the calibration dataset \(\mathcal{D}_{\text{cal}}^{\text{TG}}\).

    \item \textbf{Phase 2: Layer-Wise Computation}:
    Using the calibration dataset \(\mathcal{D}_{\text{cal}}\), forward propagation is performed to compute layer outputs \(\epsilon_\theta^{(l)}(x_t, t)\). Subsequently, gradients \(\partial \mathcal{L} / \partial \epsilon_\theta^{(l)}\) are calculated through backward propagation.

    \item \textbf{Phase 3: Time-Aware Quantization}:
    For CNN and linear layers, let \(\Delta_X\) and \(\Delta_W \) be the quantization parameters of activations and weights, respectively. 
    These parameters are alternately optimized using HO over \(R\) iterations to minimize the quantization error defined in \eqref{final_obj}.
    If the layer is the post-GELU, \(\Delta_X\) is updated using MRQ; otherwise, uniform quantization is applied.
    For MatMul layers, let \(\Delta_A \) and \(\Delta_B \) be the input activations. If the layer is the post-softmax, \(\Delta_A^{l,\mathcal{G}_i}\) is updated using time-grouping quantization (TGQ) combined with MRQ; otherwise, uniform quantization is applied.
\end{itemize}

The overall procedure is summarized in Algorithm \ref{alg}.

\section{Experiments}
\label{sec:experiments}

\subsection{Experimental Settings}
We evaluate the proposed TQ-DiT with three metrics. Fréchet Inception Distance (FID) \cite{FID} measures the similarity between feature distributions of generated and real images, where lower values indicate higher fidelity. Spatial FID (sFID) \cite{sFID}, an extension of FID, evaluates the spatial coherence of the generated images by comparing spatial feature distributions. Inception Score (IS) \cite{IS} assesses the quality and diversity of generated images, with higher scores reflecting better performance.

We experiment on the ImageNet~\cite{ImageNet} dataset using the DiT-XL-2 model \cite{DiT}, generating 10,000 images at a resolution of $256\times256$ (10 samples per class) for evaluation.
DDPM is implemented for the diffusion process with timesteps $T=100$ and $T=250$. Both weights (W) and activations (A) are quantized with 8-bit precision (W8A8) and 6-bit precision (W6A6). The iteration round $R$ is set to 3. We employ $32$ calibration samples per group, divided across $10$ timestep groups, ensuring the same number of calibration samples for all baseline schemes.

The performance of the proposed TQ-DiT is evaluated against three quantization schemes: Q-Diffusion \cite{Q-Diffusion}, PTQD \cite{PTQD}, and PTQ4DiT \cite{PTQ4DiT}. Q-Diffusion and PTQD originally have applied PTQ to diffusion models, and for comparison, they are implemented on DiT. Additionally, we compare TQ-DiT with PTQ4DiT, which has demonstrated remarkable results in PTQ for DiTs. The full-precision model (32 bit) is referred to as FP.

\begin{table}[t]
\caption{Performance comparison for timesteps of 250 on ImageNet 256×256. ‘(W/A)’ indicates that the precision of weights and activations are W and A bits, respectively.}
\label{tab1}
\centering
\begin{tabular}{cccccc}
\toprule
Bit                  & Method              & FID (↓) & sFID (↓) & IS (↑) \\ \midrule
\color{gray}
32/32                   & \color{gray}FP      & \color{gray}4.62    & \color{gray}18.00    & \color{gray}190.61 \\ \midrule
\multirow{4}{*}{8/8} & Q-Diffusion~\cite{Q-Diffusion} & 6.89    & 22.34    & 167.82 \\
                     & PTQD~\cite{PTQD}               & 6.21    & 20.16    & 169.13 \\
                     & PTQ4DiT~\cite{PTQ4DiT}         & 5.85    & 19.38    & 171.05 \\
                     & \textbf{TQ-DiT (Ours)}                  & \textbf{4.91}& \textbf{18.42}& \textbf{187.22} \\ \midrule
\multirow{4}{*}{6/6} & Q-Diffusion~\cite{Q-Diffusion} & 28.86   & 34.93    & 73.43  \\
                     & PTQD~\cite{PTQD}               & 17.59   & 27.68    & 125.59  \\
                     & PTQ4DiT~\cite{PTQ4DiT}         & 20.53   & 32.60    & 73.95 \\
                     & \textbf{TQ-DiT (Ours)}                  & \textbf{8.58}    & \textbf{27.22}    & \textbf{156.21} \\ 
                     \bottomrule
\end{tabular}
\end{table}

\begin{table}[t]
\caption{Performance comparison for timesteps of 100 on ImageNet 256×256.}
\label{tab2}
\centering
\begin{tabular}{cccccc}
\toprule
Bit                  & Method              & FID (↓) & sFID (↓) & IS (↑) \\ \midrule
\color{gray}32/32    & \color{gray}FP      &\color{gray}4.87 &\color{gray}18.78 &\color{gray}184.59 \\ \midrule
\multirow{4}{*}{8/8} & Q-Diffusion~\cite{Q-Diffusion} & 6.06    & 19.29    & 170.96 \\
                     & PTQD~\cite{PTQD}               & 6.02    & 20.50    & 187.53 \\
                     & PTQ4DiT~\cite{PTQ4DiT}         & 6.01    & 19.40    & 172.62 \\
                     & \textbf{TQ-DiT (Ours)}         & \textbf{5.10}& \textbf{18.88}& \textbf{188.45} \\ \midrule
\multirow{4}{*}{6/6} & Q-Diffusion~\cite{Q-Diffusion} & 37.81   & 35.47    & 66.86  \\
                     & PTQD~\cite{PTQD}               & 19.12   & 39.45    & 105.16 \\
                     & PTQ4DiT~\cite{PTQ4DiT}         & 24.35   & \textbf{22.55}    & 64.78  \\
                     & \textbf{TQ-DiT (Ours)} & \textbf{15.97}   & 28.37    & \textbf{108.73} \\ 
                     \bottomrule
\end{tabular}
\end{table}

\subsection{Performance Comparison}

As shown in Table \ref{tab1}, at W8A8 with 250 timesteps, TQ-DiT achieves FID of 4.91, maintaining the highest IS of 187.22. Similarly, at W8A8 with 100 timesteps, it achieves FID of 5.10 and IS of 188.45, as shown in Table \ref{tab2}.
In the more challenging W6A6 setting, the performance gap between methods is more noticeable. At 250 timesteps, baseline methods exhibit substantial degradation, with FID increasing considerably. In contrast, TQ-DiT limits the FID increase to 8.58, considerably lower than other methods. Similarly, at 100 timesteps, TQ-DiT achieves FID of 15.97 and the highest IS of 108.73.
Fig. \ref{fig1} illustrates the overall performance comparison with both W8A8 and W6A6 over 250 sampling steps. TQ-DiT outperforms other quantization schemes and achieves performance close to the FP model. While the three baseline methods suffer significant degradation in lower-bit setting (W6A6), TQ-DiT effectively preserves image quality, maintaining strong performance even under aggressive quantization.

Fig. \ref{fig5} demonstrates a visual comparison of generated images with the state-of-the-art PTQ4DiT \cite{PTQ4DiT} to clearly observe the overall outputs.
Under the W8A8 condition, both TQ-DiT and PTQ4DiT generate clearly distinguishable images. However, TQ-DiT produces sharper and more visually refined results. 
Furthermore, under the stricter W6A6 condition, PTQ4DiT tends to generate simpler images, whereas TQ-DiT excels at preserving fine details.

\begin{table}[t]
\caption{Ablation study on ImageNet 256×256 with W6A6.}
\label{tab3}
\begin{center}
\begin{tabular}{cccccc}
\toprule
Method      & FID (↓) & sFID (↓) & IS(↑) \\
\midrule
\color{gray}FP          & \color{gray}4.62    & \color{gray}18.00    & \color{gray}190.61  \\ 
\midrule
Baseline & 28.86   & 34.93    & 73.43                 \\
$+$ HO     & 22.47   & 31.65    & 89.21                \\
$+$ HO $+$ MRQ     & 9.31   & 28.34    & 143.68                \\
\textbf{$+$ HO $+$ MRQ $+$ TGQ} & \textbf{8.58} & \textbf{27.22}& \textbf{156.21}    \\ 
\bottomrule
\end{tabular}
\end{center}
\end{table}

\subsection{Ablation Study}
We investigate the impact of HO, MRQ, and TGQ on DiTs through an ablation study under the W6A6 condition. Four configurations are considered: \textbf{(a) Baseline} applies standard uniform quantization with MSE-based optimization for DiTs. \textbf{(b) Baseline + HO} employs HO for optimization. \textbf{(c) Baseline + HO + MRQ} extends (b) by incorporating MRQ to quantize activations in the MHSA and PF modules. \textbf{(d) Baseline + HO + MRQ + TGQ} represents the complete TQ-DiT framework, fully integrating TGQ. 
Table \ref{tab3} demonstrates that each component enhances performance. 
In particular, applying HO and MRQ delivers notable improvements over the
baseline, reducing FID by 6.39 and 13.16, and sFID by 3.28 and 3.31, respectively.
Moreover, the integration of TGQ further elevates the performance of TQ-DiT, emphasizing its crucial role in addressing time-dependent variations.

\subsection{Computational Efficiency}
\begin{table}[t]
\caption{Efficiency comparison of calibration algorithm.}
\label{tab4}
\begin{center}
\begin{tabular}{ccc}
\toprule
Method                 & GPU memory (GB) & GPU times (hour) \\ \midrule
PTQ4DiT~\cite{PTQ4DiT} & 1.59            & 11.54            \\
\textbf{TQ-DiT (Ours)} & \textbf{0.87}   & \textbf{1.23}    \\ \midrule
Reduction (\%)         & 45.4\% lower    & 89.3\% lower     \\ \bottomrule
\end{tabular}
\end{center}
\end{table}

Computational efficiency of TQ-DiT is evaluated through a comparison with PTQ4DiT \cite{PTQ4DiT}.
Although both methods use identical bit-widths, so that memory usage of parameters for DiT remains the same, there is a noteworthy difference in calibration overhead. 
As shown in Table \ref{tab4}, TQ-DiT employs 45.4\% less GPU memory than PTQ4DiT during calibration. 
Additionally, TQ-DiT reduces calibration time by 89.3\% relative to PTQ4DiT.
Thus, TQ-DiT not only improves computational efficiency but also aligns with the goals of sustainable AI by significantly reducing GPU resource consumption. 
This efficiency makes TQ-DiT practical for deployment in resource-constrained environments, contributing to greener and more accessible generative AI systems.


\section{Conclusion}
\label{sec:conclusion}
In this paper, we propose TQ-DiT, a novel quantization framework for diffusion transformers that mitigates activation imbalance and temporal variability in generative processes. 
Through a time-aware quantization algorithm, TQ-DiT ensures robust performance under low-bit settings while preserving generative quality and reducing computational and memory overhead, making it practical for resource-constrained environments.
Furthermore, by significantly lowering GPU memory usage and demanding the time required for calibration, TQ-DiT promotes more sustainable AI practices, reducing the energy footprint of post-training quantization for diffusion-based generative models.
The success of TQ-DiT underscores the importance of accounting for temporal dynamics and architectural intricacies in generative models, with potential applications beyond diffusion transformers to other temporal and sequential architectures. 
Future work includes extending our approach to video generation and multi-modal learning, as well as developing adaptive calibration techniques for real-time applications.

\bibliographystyle{IEEEtran}
\bibliography{refs}

\begin{thebibliography}{10}
\providecommand{\url}[1]{#1}
\csname url@samestyle\endcsname
\providecommand{\newblock}{\relax}
\providecommand{\bibinfo}[2]{#2}
\providecommand{\BIBentrySTDinterwordspacing}{\spaceskip=0pt\relax}
\providecommand{\BIBentryALTinterwordstretchfactor}{4}
\providecommand{\BIBentryALTinterwordspacing}{\spaceskip=\fontdimen2\font plus
\BIBentryALTinterwordstretchfactor\fontdimen3\font minus \fontdimen4\font\relax}
\providecommand{\BIBforeignlanguage}[2]{{%
\expandafter\ifx\csname l@#1\endcsname\relax
\typeout{** WARNING: IEEEtran.bst: No hyphenation pattern has been}%
\typeout{** loaded for the language `#1'. Using the pattern for}%
\typeout{** the default language instead.}%
\else
\language=\csname l@#1\endcsname
\fi
#2}}
\providecommand{\BIBdecl}{\relax}
\BIBdecl

\bibitem{DMsurvey2}
L.~Yang, Z.~Zhang, Y.~Song, S.~Hong, R.~Xu, Y.~Zhao, W.~Zhang, B.~Cui, and M.-H. Yang, ``Diffusion models: A comprehensive survey of methods and applications,'' \emph{ACM Comput. Surv.}, vol.~56, no.~4, pp. 1--39, 2023.

\bibitem{ADM}
P.~Dhariwal and A.~Nichol, ``Diffusion models beat gans on image synthesis,'' in \emph{Proc. NeurIPS}, Virtual Event, Dec. 2021.

\bibitem{LDM}
R.~Rombach, A.~Blattmann, D.~Lorenz, P.~Esser, and B.~Ommer, ``High-resolution image synthesis with latent diffusion models,'' in \emph{Proc. IEEE/CVF CVPR}, New Orleans, US, June 2022.

\bibitem{ViT}
A.~Dosovitskiy, L.~Beyer, A.~Kolesnikov, D.~Weissenborn, X.~Zhai, T.~Unterthiner, M.~Dehghani, M.~Minderer, G.~Heigold, S.~Gelly, J.~Uszkoreit, and N.~Houlsby, ``An image is worth 16x16 words: Transformers for image recognition at scale,'' in \emph{Proc. ICLR}, Vienna, Austria, May 2021.

\bibitem{Transformer1}
N.~Carion, F.~Massa, G.~Synnaeve, N.~Usunier, A.~Kirillov, and S.~Zagoruyko, ``End-to-end object detection with transformers,'' in \emph{Proc. ECCV}, Virtual Event, Aug. 2020.

\bibitem{DiT}
W.~Peebles and S.~Xie, ``Scalable diffusion models with transformers,'' in \emph{Proc. ICCV}, Paris, France, Oct. 2023.

\bibitem{DiT2}
T.~Brooks, B.~Peebles, C.~Holmes, W.~DePue, Y.~Guo, L.~Jing, D.~Schnurr, J.~Taylor, T.~Luhman, E.~Luhman, C.~Ng, R.~Wang, and A.~Ramesh, ``Video generation models as world simulators,'' 2024.

\bibitem{DiT3}
S.~Gao, P.~Zhou, M.-M. Cheng, and S.~Yan, ``Masked diffusion transformer is a strong image synthesizer,'' in \emph{Proc. ICCV}, Paris, France, Oct. 2023.

\bibitem{DiT4}
J.~Chen, J.~Yu, C.~Ge, L.~Yao, E.~Xie, Y.~Wu \emph{et~al.}, ``Pixart-alpha: Fast training of diffusion transformer for photorealistic text-to-image synthesis,'' \emph{arXiv preprint arXiv:2310.00426}, 2023.

\bibitem{SusAI1}
V.~Bol{\'o}n-Canedo, L.~Mor{\'a}n-Fern{\'a}ndez, B.~Cancela, and A.~Alonso-Betanzos, ``A review of green artificial intelligence: Towards a more sustainable future,'' \emph{Neurocomputing}, vol. 599, p. 128096, 2024.

\bibitem{SusAI2}
C.-J. Wu, R.~Raghavendra, U.~Gupta, B.~Acun, N.~Ardalani, K.~Maeng, G.~Chang, F.~Aga, J.~Huang, C.~Bai \emph{et~al.}, ``Sustainable ai: Environmental implications, challenges and opportunities,'' in \emph{Proc. MLSys}, Santa Clara, US, Aug. 2022.

\bibitem{Qinference}
B.~Jacob, S.~Kligys, B.~Chen, M.~Zhu, M.~Tang, A.~Howard, H.~Adam, and D.~Kalenichenko, ``Quantization and training of neural networks for efficient integer-arithmetic-only inference,'' in \emph{Proc. IEEE/CVF CVPR}, Salt Lake City, US, June 2018.

\bibitem{PTQ1}
Y.~Li, R.~Gong, X.~Tan, Y.~Yang, P.~Hu, Q.~Zhang, F.~Yu, W.~Wang, and S.~Gu, ``Brecq: Pushing the limit of post-training quantization by block reconstruction,'' in \emph{Proc. ICLR}, Virtual Event, May 2021.

\bibitem{PTQ2}
M.~Nagel, R.~A. Amjad, M.~Van~Baalen, C.~Louizos, and T.~Blankevoort, ``Up or down? adaptive rounding for post-training quantization,'' in \emph{Proc. ICML}, Virtual Event, July 2020.

\bibitem{PTQ3}
Z.~Yuan, C.~Xue, Y.~Chen, Q.~Wu, and G.~Sun, ``Ptq4vit: Post-training quantization for vision transformers with twin uniform quantization,'' in \emph{Proc. ECCV}, Tel Aviv, Israel, Oct. 2022.

\bibitem{PTQ4DiT}
J.~Wu, H.~Wang, Y.~Shang, M.~Shah, and Y.~Yan, ``Ptq4dit: Post-training quantization for diffusion transformers,'' \emph{arXiv preprint arXiv:2405.16005}, 2024.

\bibitem{energy}
F.~Ayaz and M.~Nekovee, ``Ai-based energy consumption modeling of 5g base stations: an energy efficient approach,'' in \emph{IET 6G 2024}, London, UK, June 2024.

\bibitem{Semantic}
E.~Erdemir, T.-Y. Tung, P.~L. Dragotti, and D.~Gündüz, ``Generative joint source-channel coding for semantic image transmission,'' \emph{JSAC}, vol.~41, no.~8, pp. 2645--2657, 2023.

\bibitem{Semantic2}
M.~Nekovee and F.~Ayaz, ``Vision, enabling technologies, and scenarios for a 6g-enabled internet of verticals (6g-iov),'' \emph{Future Internet}, vol.~15, no.~2, p.~57, 2023.

\bibitem{DM}
J.~Ho, A.~Jain, and P.~Abbeel, ``Denoising diffusion probabilistic models,'' in \emph{Proc. NeurIPS}, Virtual Event, Dec. 2020.

\bibitem{PTQ4}
M.~Nagel, M.~Fournarakis, R.~A. Amjad, Y.~Bondarenko, M.~Van~Baalen, and T.~Blankevoort, ``A white paper on neural network quantization,'' \emph{arXiv preprint arXiv:2106.08295}, 2021.

\bibitem{PTQD}
Y.~He, L.~Liu, J.~Liu, W.~Wu, H.~Zhou, and B.~Zhuang, ``Ptqd: Accurate post-training quantization for diffusion models,'' in \emph{Proc. NeurIPS}, New Orleans, US, Dec. 2023.

\bibitem{QLLM}
G.~Xiao, J.~Lin, M.~Seznec, H.~Wu, J.~Demouth, and S.~Han, ``Smoothquant: Accurate and efficient post-training quantization for large language models,'' in \emph{Proc. ICML}, Honolulu, US, July 2023.

\bibitem{QLLM2}
T.~Dettmers, M.~Lewis, Y.~Belkada, and L.~Zettlemoyer, ``Gpt3.int8(): 8-bit matrix multiplication for transformers at scale,'' in \emph{Proc. NeurIPS}, New Orleans, US, Dec. 2022.

\bibitem{Q-Diffusion}
X.~Li, Y.~Liu, L.~Lian, H.~Yang, Z.~Dong, D.~Kang, S.~Zhang, and K.~Keutzer, ``Q-diffusion: Quantizing diffusion models,'' in \emph{Proc. ICCV}, Paris, France, Dec. 2023.

\bibitem{chainnet_edge_ai}
Z.~Niu, M.~Roveri, and G.~Casale, ``Chainnet: A customized graph neural network model for loss-aware edge ai service deployment,'' in \emph{Proc. IEEE/IFIP DSN}, Brisbane, Australia, June 2024.

\bibitem{MSE1}
Y.~Choukroun, E.~Kravchik, F.~Yang, and P.~Kisilev, ``Low-bit quantization of neural networks for efficient inference,'' in \emph{Proc. IEEE/CVF ICCV Workshops}, Seoul, Korea, Oct. 2019.

\bibitem{MSE2}
Z.~Liu, Y.~Wang, K.~Han, W.~Zhang, S.~Ma, and W.~Gao, ``Post-training quantization for vision transformer,'' in \emph{Proc. NeurIPS}, Virtual Event, Dec. 2021.

\bibitem{FID}
M.~Heusel, H.~Ramsauer, T.~Unterthiner, B.~Nessler, and S.~Hochreiter, ``Gans trained by a two time-scale update rule converge to a local nash equilibrium,'' in \emph{Proc. NeurIPS}, Long Beach, US, Dec. 2017.

\bibitem{sFID}
C.~Nash, J.~Menick, S.~Dieleman, and P.~W. Battaglia, ``Generating images with sparse representations,'' \emph{arXiv preprint arXiv:2103.03841}, 2021.

\bibitem{IS}
T.~Salimans, I.~Goodfellow, W.~Zaremba, V.~Cheung, A.~Radford, X.~Chen, and X.~Chen, ``Improved techniques for training gans,'' in \emph{Proc. NeurIPS}, Barcelona, Spain, 2016.

\bibitem{ImageNet}
O.~Russakovsky, J.~Deng, H.~Su, J.~Krause, S.~Satheesh, S.~Ma, Z.~Huang, A.~Karpathy, A.~Khosla, M.~Bernstein \emph{et~al.}, ``Imagenet large scale visual recognition challenge,'' \emph{IJCV}, vol. 115, pp. 211--252, 2015.

\end{thebibliography}

\end{document}